\documentclass{article}

\usepackage{arxiv}

\usepackage[utf8]{inputenc} 
\usepackage[T1]{fontenc}    
\usepackage{hyperref}       
\usepackage{url}            
\usepackage{booktabs}       
\usepackage{amsfonts}       
\usepackage{nicefrac}       
\usepackage{microtype}      
\usepackage{lipsum}		
\usepackage{graphicx}
\usepackage{natbib}
\usepackage{doi}
\usepackage{amsmath}
\usepackage{amssymb}
\usepackage{multirow}
\usepackage{algorithm}
\usepackage{algcompatible}
\usepackage{amsthm}
\theoremstyle{plain}
\newtheorem{theorem}{Theorem}[section]

\theoremstyle{definition}
\newtheorem{definition}[theorem]{Definition}

\theoremstyle{remark}
\newtheorem{remark}[theorem]{Remark}

\title{Improving Role Consistency in Multi‑Agent Collaboration via Quantitative Role Clarity}

\author{ {\includegraphics[scale=0.06]{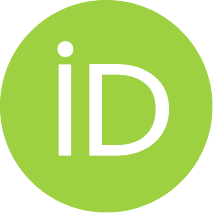}\hspace{1mm}Guoling Zhou}\\
	School of Information Science and Technology\\
	Northeast Normal University\\
	Changchun, Jilin \\
	\texttt{guolingzhou@nenu.edu.cn} \\
	\And
	{\includegraphics[scale=0.06]{orcid.pdf}\hspace{1mm}Wenpei Han} \\
	School of Information Science and Technology\\
	Northeast Normal University\\
	Changchun, Jilin \\
	\texttt{wphan@nenu.edu.cn} \\
	\AND
	{\includegraphics[scale=0.06]{orcid.pdf}\hspace{1mm}Fengqin Yang} \\
	School of Information Science and Technology\\
	Northeast Normal University \\
	Changchun, Jilin \\
	\texttt{yangfq147@nenu.edu.cn} \\
	\And
	{\includegraphics[scale=0.06]{orcid.pdf}\hspace{1mm}Li Wang} \\
	School of Computer Science and Engineering \\
	Guangxi Normal University \\
	Guilin, Guangxi\\
\texttt{wangl028@gxnu.edu.cn} \\
	\And
	{\includegraphics[scale=0.06]{orcid.pdf}\hspace{1mm}Yingcong Zhou} \\
	School of Information Science and Technology \\
	Northeast Normal University \\
    Changchun, Jilin \\
	\texttt{zhouyc821@nenu.edu.cn} \\
    \And
	{\includegraphics[scale=0.06]{orcid.pdf}\hspace{1mm}Zhiguo Fu}\thanks{Corresponding author.} \\
	School of Information Science and Technology\\
	Northeast Normal University \\
	Changchun, Jilin \\
    \texttt{fuzg432@nenu.edu.cn} \\
}

\hypersetup{
pdftitle={A template for the arxiv style},
pdfsubject={q-bio.NC, q-bio.QM},
pdfauthor={David S.~Hippocampus, Elias D.~Striatum},
pdfkeywords={First keyword, Second keyword, More},
}

\begin{document}
\maketitle

\begin{abstract}
In large language model (LLM)-driven multi-agent systems, disobey role specification (failure to adhere to the defined responsibilities and
constraints of an assigned role, potentially leading to an agent behaving like another) is a major failure mode \cite{DBLP:journals/corr/abs-2503-13657}. To address this issue, in the present paper, we propose a quantitative role clarity to improve role consistency. Firstly, we construct a role assignment matrix $S(\phi)=[s_{ij}(\phi)]$, where $s_{ij}(\phi)$ is the semantic similarity  between the $i$-th agent's  behavior trajectory and the $j$-th agent's role description. Then we define  role clarity matrix $M(\phi)$ as $\text{softmax}(S(\phi))-I$, where $\text{softmax}(S(\phi))$ is a row-wise softmax of $S(\phi)$ and $I$ is the identity matrix. The Frobenius norm of $M(\phi)$ quantifies the alignment  between agents' role descriptions  and their behaviors trajectory.
Moreover, we employ the role clarity matrix as a regularizer during lightweight fine-tuning to improve role consistency, thereby improving end-to-end task performance.
Experiments on the ChatDev multi-agent system show that our method substantially improves role consistency and task performance: with Qwen and Llama, the role overstepping rate decreases from $46.4\%$ to $8.4\%$ and from $43.4\%$ to $0.2\%$, respectively, and the role clarity score increases from $0.5328$ to $0.9097$ and from $0.5007$ to $0.8530$, respectively, the  task success rate  increases from $0.6769$ to $0.6909$ and from $0.6174$ to $0.6763$, respectively.
\end{abstract}

\section{Introduction}
Recently, LLM-driven multi-agent systems achieved great progresses in solving complex tasks   \cite{qian-etal-2024-chatdev, NEURIPS2023_a3621ee9, DBLP:conf/iclr/HuLC25, DBLP:journals/corr/abs-2503-03686, DBLP:journals/fcsc/WangMFZYZCTCLZWW24, DBLP:journals/corr/abs-2310-08560}, in particular for highly collaborative applications such as software development.
By equipping agents with distinct role profiles and interaction protocols \cite{DBLP:journals/corr/abs-2308-08155, DBLP:conf/iclr/ChenSZ0YCYLHQQC24}, 
these systems can coordinate as a role-specialized team—enabling division of labor and collective reasoning—so that long-horizon tasks can be decomposed into jointly executable sub-processes.
The well-known multi-agent systems,  such as ChatDev and MetaGPT for automated software engineering \cite{qian-etal-2024-chatdev, DBLP:conf/iclr/HongZCZCWZWYLZR24} and AutoGen for more general multi-agent tasks \cite{DBLP:conf/ieeecai/BarbarroxaRGV24}, all leverage structured collaboration to improve reliability and end-to-end completion of complex tasks \cite{DBLP:conf/icml/Du00TM24, DBLP:conf/emnlp/Liang0JW00Y0T24, DBLP:conf/iclr/ChenSZ0YCYLHQQC24, DBLP:conf/iclr/ZhangXYTCCZCHWZ25, liu2024dynamicllmpoweredagentnetwork, DBLP:conf/icml/ZhugeWKFKS24}.

However, empirical studies reveal that the multi-agent systems are often unstable  \cite{DBLP:journals/corr/abs-2407-01489, DBLP:journals/tmlr/KapoorSSNN25, DBLP:conf/acl/QianLLCDL0CSCXL24}. One major source of this instability is role inconsistency during collaborative interactions,  i.e., an agent fails to adhere to the defined responsibilities and
constraints of an assigned role, potentially leading to an agent behaving like another,ultimately reducing end-to-end task success rates substantially \cite{DBLP:journals/corr/abs-2503-13657}.

In  social psychology and organizational behavior, there is a well-known phenomenon: role ambiguity and role conflict significantly reduce team performance and collaboration efficiency, whereas role clarity is positively associated with higher performance and consistency \cite{TUBRE2000155, HALL2008141}. Inspired by this insight, we define the role clarity for multi-agent collaboration. Because current multi‑agent systems primarily define roles through natural-language prompts, which is not quantifiable and differentiable. Thus a  definition of role clarity should be:

\textbf{quantifiable:} it can measure the alignment between the role descriptions  and  the behavior trajectories, and measure the overstepping from an agent's behavior to other roles' responsibility boundaries, to  reduce agents’ prompt sensitivity in prompt engineering \cite{DBLP:journals/corr/abs-2411-10541, DBLP:journals/corr/abs-2306-03314, DBLP:conf/iclr/ChanCSYXZF024, wu2023autogenenablingnextgenllm}.

\textbf{differentiable:} it can be incorporated as a training constraint (e.g., regularizer) to correct behavior and internalize the role consistency through gradient-based optimization, 
reducing reliance on reinforcement learning with carefully engineered reward functions \cite{DBLP:conf/nips/YuVVGWBW22, guo2024heterogeneousmultiagentreinforcementlearning, DBLP:journals/corr/abs-2410-08115}.

\begin{figure*}[t]
\centering
\includegraphics[width=1\textwidth]{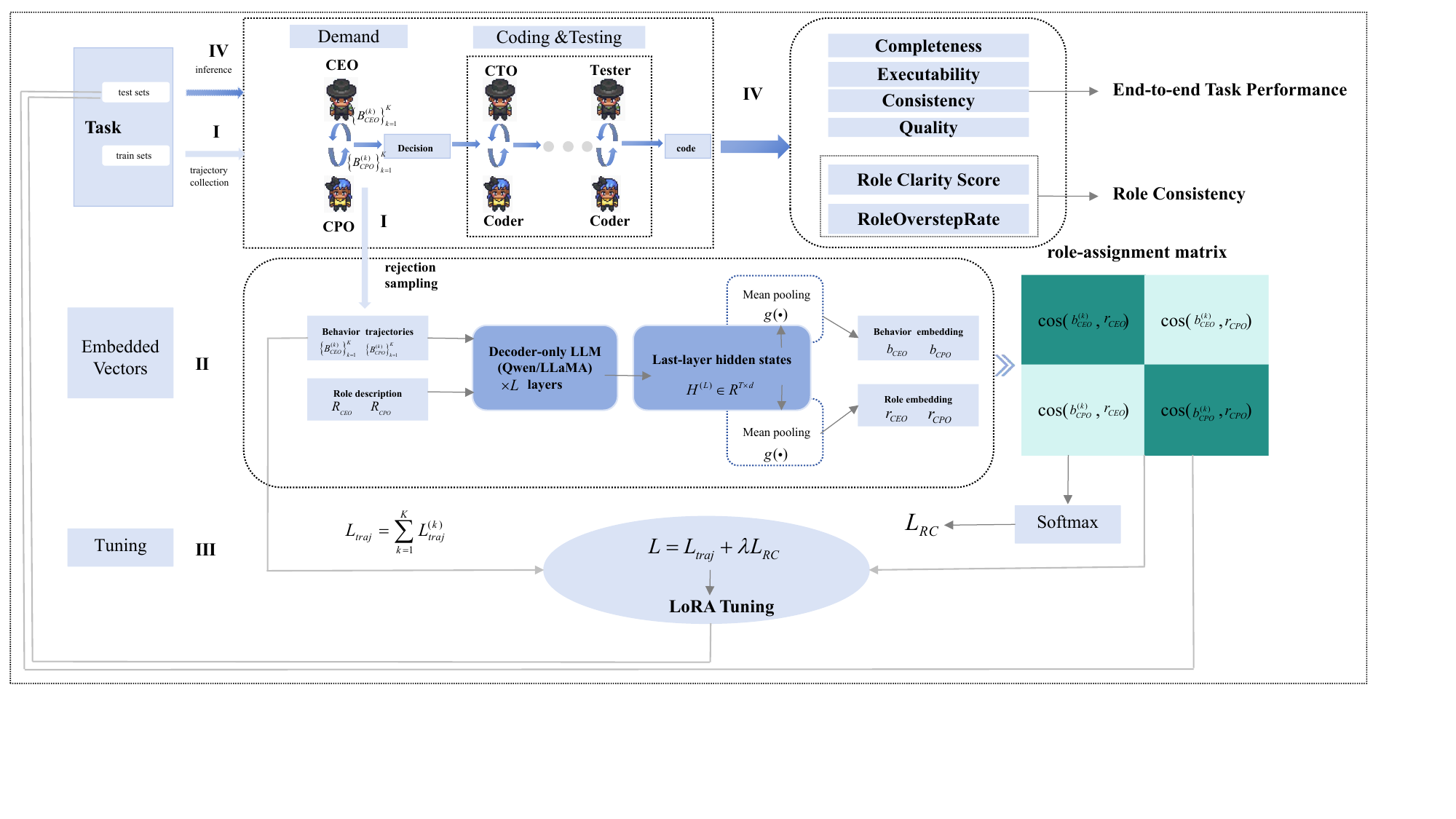} 
\caption{
{\bf Overview of the LoRA‑tuning framework with role clarity regularization.}
 The framework comprises four stages: (I) collecting high‑quality multi‑turn interaction trajectories via rejection sampling, (II) extracting embeddings and computing  role assignment matrix based on similarity, (III) role clarity‑regularized fine tuning using LoRA, and (IV) evaluating role consistency and end-to-end task performance in  multi‑agent interactions.
 }
\label{p1}
\end{figure*}

In the present paper, we propose a quantifiable and differentiable role clarity 
and employ the role clarity as a regularizer to improve role consistency via lightweight fine tuning and enhance end-to-end task success rate. An overview of this framework is presented in Figure ~\ref{p1}. 
Our main contributions are as follows.

\begin{figure*}[t]
\centering
\includegraphics[width=1\textwidth]{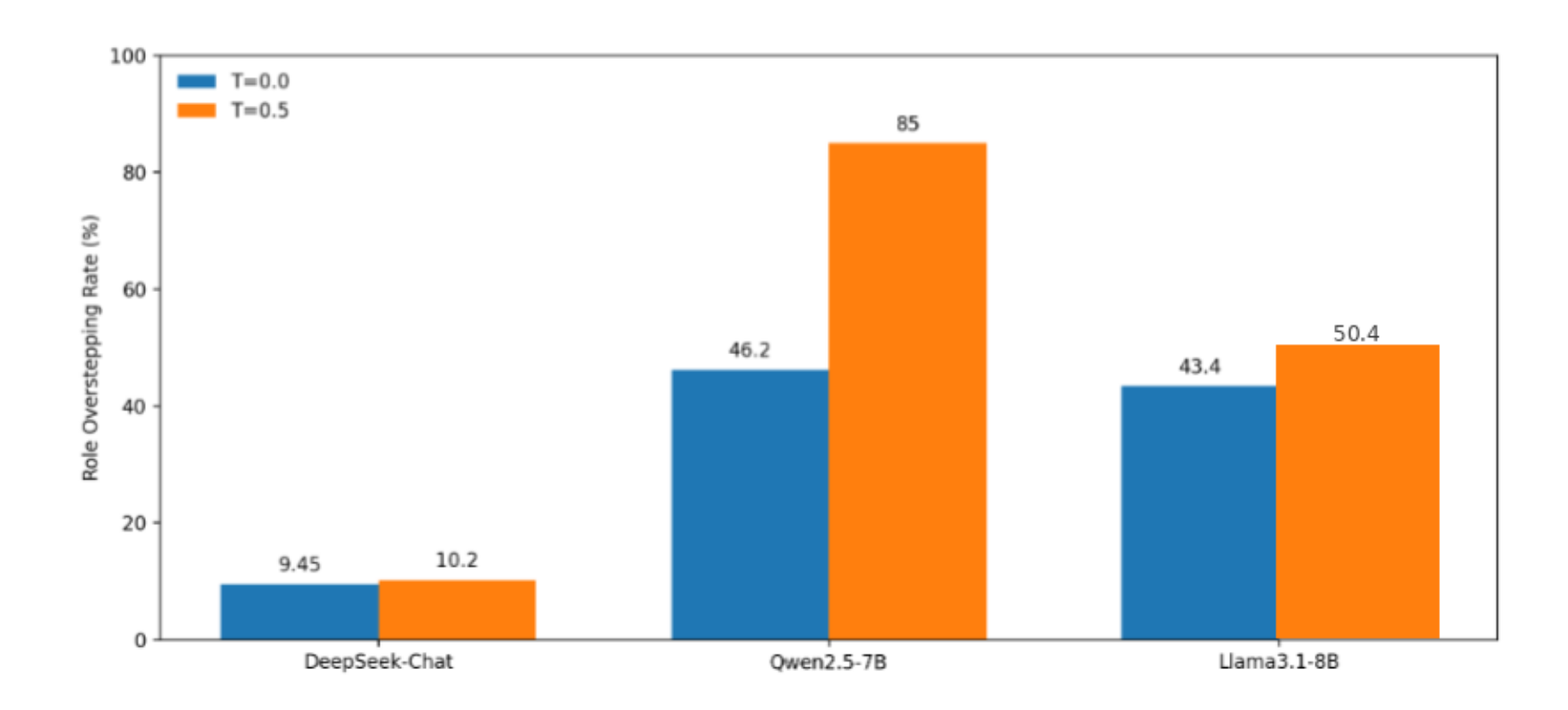} 
\caption{
{\bf Role overstepping rates for DeepSeek Chat, Qwen2.5 7B, and Llama3.1 8B on SWE‑Dev under ChatDev.}
Lower temperature $T$ yields more deterministic output, whereas higher temperature increases diversity.
 }
\label{historical}
\end{figure*}

\textbf{Evaluating Role Overstepping in Multi-Agent Systems.} We conduct experiments for  the ChatDev framework on the mainstream language models (Figure ~\ref{historical}). The results show that role inconsistency is a key factor underlying failures in current multi‑agent systems.

\textbf{Proposing a Quantifiable Role Clarity.} 
From the perspective of role consistency, we  propose a definition of role clarity that satisfies global deviation measurability, monotonic consistency, and continuous differentiability, providing an optimizable guarantee for structurally constraining agent behavior. To our best knowledge, this is the first definition of quantitative role clarity.

\textbf{Constructing Dataset.} We apply rejection sampling based on the termination token \texttt{<INFO>} (i.e., no overstepping) to collect multi-turn behavior trajectories to fine tuning agents.

\textbf{Developing a Lightweight Fine Tuning Framework.} We incorporate role clarity as a regularizer in the optimization objective, guiding agents  to adherence to role responsibilities and improving role consistency in multi-agent systems.

\textbf{Empirical Validation on the ChatDev Benchmark.} Experimental results demonstrate that optimizing role clarity can significantly reduce the role overstepping and improve role consistency.

\textbf{Improving End-to-End Task performance.}
Experimental results demonstrate that our method effectively improves end-to-end task performance on the downstream SRDD (Software
Requirement Description Dataset)
benchmark.

\section{Quantifiable Role Clarity}
In this section, we propose a quantifiable definition of role clarity for multi-agent collaboration. We begin by introducing an abstract agent representation, and then define a role clarity  that captures (i) each agent’s adherence to its assigned role and (ii) its overlap with other agents’ roles.
\subsection{Agent Representation}
We consider a multi-agent system with $n$ LLM-based agents $\{A_i\}_{i=1}^n$. 
Each agent $A_i$ is represented as a tuple
$A_i = (L_i, R_i, B_i),$
where\\
\textbf{(i)}~$L_i$ denotes the underlying language model instance (architecture and inference configuration) used by agent $A_i$;\\
\textbf{(ii)}~$R_i$ is a textual role description by natural language, including the intended responsibilities and boundary of $A_i$ (e.g., ``Chief Executive Officer (CEO)'' or ``Chief Product Officer (CPO)'' with task-specific expectations);\\
\textbf{(iii)}~$B_i$ is the behavior trajectory of $A_i$ during interaction.

Based on the  agent representation above, a collaboration requires each agent to adhere to its assigned role: $B_i$ should align with $R_i$ and avoid overstepping other agents' responsibilities. However, role descriptions are  specified in natural language and rarely provide sharp quantitative boundaries, making it challenging to measure (i) the misalignment between $B_i$ and $R_i$, and (ii) the overstepping from $B_i$ to the responsibilities of role $R_j$ $(j\neq i)$.
Moreover, prompt-based role constraints are ``soft" and degrade in long-horizon, multi-round interactions due to context limitations.

These challenges motivate a quantifiable role clarity that simultaneously captures role consistency and overstepping.
\subsection{Definition of the Role Clarity}
Consider $n$ agents $\{A_i=(L_i,R_i,B_i)\}_{i=1}^n$ collaboratively solving a task $\mathcal{T}$. 
Over $K$ interaction rounds, the resulting behavior trajectory is denoted by $\{B_i^{(k)}\}_{k=1}^{K}$.
To obtain a quantifiable role clarity, we follow a three-step procedure: 
\paragraph{(i) Role descriptions and behavior trajectories embedding vectors.}\label{subsec:rc_def}
We embed both role descriptions and behavior trajectories into a shared $d$-dimensional semantic space using an encoder $f_{\phi}$.
Let $\mathbf{h}_{R_i,1:M}(\phi)$ and $\mathbf{h}_{\{B_i^{(k)}\}_{k=1}^{K},1:T}(\phi)$ denote  hidden states produced  by the final-layer of $f_{\phi}$ for the role description $R_i$ (length $M$) and the behavior trajectory $\{B_i^{(k)}\}_{k=1}^{K}$ (length $T$), respectively.
We then apply an aggregation (mean pooling) operator $\mathcal{A}$ to obtain sequence-level embeddings:
\begin{align*}
\mathbf{r}_i(\phi) = \mathcal{A}\!\left(\mathbf{h}_{R_i,1:M}(\phi)\right), ~
\mathbf{b}_i(\phi) = \mathcal{A}\!\left(\mathbf{h}_{\{B_i^{(k)}\}_{k=1}^{K},1:T}(\phi)\right).
\end{align*}
Notably, during fine tuning, the role embeddings \(\{\mathbf{r}_i(\phi)\}_{i=1}^n\) are obtained with the pretrained parameters fixed, whereas the behavior embeddings \(\{\mathbf{b}_i(\phi)\}_{i=1}^{n}\) are obtained with parameters that are updated through training.
\paragraph{(ii) Role clarity matrix.}
Based on the role embeddings \(\{\mathbf{r}_i(\phi)\}_{i=1}^n\) and behavior embeddings \(\{\mathbf{b}_i(\phi)\}_{i=1}^n\), we construct a role assignment matrix \(S(\phi) = [s_{ij}(\phi)]_{i,j=1}^n\) using cosine similarity:
\begin{equation*}
s_{ij}(\phi) = \cos\!\big(\mathbf{b}_i(\phi), \mathbf{r}_j(\phi)\big) \in [-1, 1],
\end{equation*}
where \(s_{ij}(\phi)\) represents the semantic similarity between agent \(A_i\)'s behavior trajectory and agent \(A_j\)'s role description. 
We then apply a row‑wise softmax with temperature \(\tau\) to \(S(\phi)\), yielding a row-stochastic role assignment matrix 
$\text{softmax}_{\tau}\big(S(\phi)\big)$:
\begin{equation*}
\text{softmax}_{\tau}\left(s_{ij}(\phi)\right) 
= \frac{\exp\!\big(s_{ij}(\phi)/\tau\big)}{\sum_{j=1}^{n}\exp\!\big(s_{ij}(\phi)/\tau\big)} 
\;\in\; (0,1),
\end{equation*}
such that \(\sum_{j=1}^{n}\text{softmax}\left(s_{ij}(\phi)\right) = 1\) for all $i$.

In the resulting matrix $\text{softmax}_{\tau}\big(S(\phi)\big)$, the diagonal entries \(\text{softmax}\left(s_{ii}(\phi)\right) \) correspond to role consistency, i.e., agent \(A_i\)'s behavior aligns with its assigned role, and  the off‑diagonal entries \(\text{softmax}_{\tau}\left(s_{ij}(\phi)\right) \) for \(i \neq j\) represent role overstepping, i.e., from agent \(A_i\)'s behavior to  agent $A_j$'s role description. When each agent strictly adheres to its assigned role without  overstepping, \(\text{softmax}_{\tau}\big(S(\phi)\big)\) is the identity matrix \(I \in \mathbb{R}^{n\times n}\).
Finally, we obtain the role clarity matrix \(M(\phi)\)
\begin{equation}\label{role}
M(\phi) = \text{softmax}_{\tau}(S(\phi)) - I \in \mathbb{R}^{n \times n}.
\end{equation}
\(M(\phi)\)  measures the deviation of $\text{softmax}_{\tau}(S(\phi))$ from  $I$, mitigates the ambiguity of natural language in prompt engineering and  provides a structured, quantitative representation of role clarity.
\paragraph{(iii) Role clarity.}
we  define role clarity via the Frobenius norm of $M(\phi)$ in the following.
\begin{definition}[$\epsilon$-role clear]\label{def:rc}
For the role clarity matrix $M(\phi)$,
if there exists $\epsilon>0$ such that
\begin{equation}
\|M(\phi)\|_{F}\le \epsilon,
\end{equation}
then the system  is said to be $\epsilon$-role clear, where $\|\cdot\|_F$ is the Frobenius norm.
\end{definition}
\begin{remark}
Definition~\ref{def:rc} formalizes role clarity as a optimizable measure of role consistency in multi-agent interactions. In particular:

$\bullet$ \textbf{Global Deviation Measure.}
The Frobenius norm $\|M(\phi)\|_F$  as a global metric jointly captures (i) role inconsistency and (ii) cross-role overstepping:
\begin{align}\label{remark}
\|M(\phi)\|_F^2=&\sum_{i\ne j}\left|\text{softmax}_{\tau}\left(s_{ij}(\phi)\right)\right|^2\nonumber\\
&+\sum_{i}\left|\text{softmax}_{\tau}\left(s_{ii}(\phi)\right)-1\right|^2,
\end{align}
where the first term aggregates cross-role overstepping, and the second term quantifies each agent's behavior deviating from compliance with its own assignment.

$\bullet$ \textbf{Monotonic Consistency.} 
 Based on (\ref{remark}), we have
\begin{align*}
\|M(\phi)\|_F^2
\le& \sum_{i}\Big(\sum_{i\ne j}\text{softmax}_{\tau}\left(s_{ij}(\phi)\right)\Big)^2\\
&+\sum_{i}\left|\text{softmax}_{\tau}\left(s_{ii}(\phi)\right)-1\right|^2\\
=&2\sum_{i}\left|\text{softmax}_{\tau}\left(s_{ii}(\phi)\right)-1\right|^2,
\end{align*}
since \(\sum_{j=1}^{n}\text{softmax}_{\tau}(s_{ij}(\phi))=1\). 
Then, increasing role consistency $\text{softmax}_{\tau}\left(s_{ii}(\phi)\right)$  reduces the overstepping probability mass 
$1-\text{softmax}_{\tau}\left(s_{ii}(\phi)\right)$ and monotonically reduces an upper bound on $\|M(\phi)\|_F$; conversely, decreasing $\text{softmax}_{\tau}\left(s_{ii}(\phi)\right)$  monotonically increases  this upper bound on $\|M(\phi)\|_F$. Therefore, $\|M(\phi)\|_F$ is monotonically
non-increasing with role consistency and non-decreasing with role overstepping, establishing a monotonic relationship with role clarity.

$\bullet$ \textbf{Differentiability.} 
Since $\text{softmax}\big(S(\phi)\big)$ is row‑wise softmax  and the identity matrix $I$ corresponds to a one‑hot assignment. Therefore, the Frobenius-based role clarity  admits a  cross-entropy surrogate: for row \(i\), the one-hot cross entropy reduces to the negative log-likelihood of the diagonal probability, i.e., 
$$-\log \text{softmax}_{\tau}(s_{ii}(\phi)),$$
which penalizes deviations from one‑hot role identities. Incorporating this term into the overall loss yields an objective that both reflects the Frobenius‑norm criterion and is compatible with standard gradient‑based optimization.
\end{remark}
Finally, we define a role clarity score to quantify the role  consistency during multi‑agent collaboration:
\begin{equation}\label{2.1}
C(M(\phi))=\frac{1}{1+\|M(\phi)\|_F}.
\end{equation}
  This score provides a compact and interpretable metric for evaluating role consistency in downstream tasks.

\section{LoRA Tuning Based on the Role Clarity Regularization}
In this section, 
we first review the  autoregressive language modeling, then construct a role-clarity regularizer, adopt the parameter-efficient Low-Rank Adaptation (LoRA) method to fine tune the model and improve role consistency.
\subsection{ Preliminaries of Autoregressive Language Models } 
\label{subsec:lm_prelim}
Let $\mathcal{D}=\{(x^{(i)},y^{(i)})\}_{i=1}^{m}$ be a training set of $m$ examples, where
$x^{(i)}$ denotes the input prompt (or context) and
$y^{(i)}=(y^{(i)}_{1},\ldots,y^{(i)}_{T_i})$ denotes the corresponding target token sequence of length $T_i$
(with tokens taking values in a fixed vocabulary).

Let $p_\phi(y^{(i)}|x^{(i)})$ denote the conditional distribution over target sequences $y^{(i)}$ given context $x^{(i)}$, as induced by a Transformer‑based autoregressive model parameterized by $\phi$. By the standard autoregressive factorization, the probability of a target sequence $y=(y^{(i)}_1,\ldots,y^{(i)}_T)$ as
\begin{equation}\nonumber
p_{\phi}(y^{(i)}|x^{(i)})
=\prod_{t=1}^{T} p_{\phi}\!\left(y^{(i)}_t| x^{(i)}, y^{(i)}_{<t}\right),
\end{equation}
where $y^{(i)}_{<t}=(y^{(i)}_1,\ldots,y^{(i)}_{t-1})$ denotes the prefix preceding position $t$ (and $y^{(i)}_{<1}$ is the empty prefix).

Under the autoregressive factorization, model training is performed via maximum likelihood estimation (MLE), i.e.,
minimizing the negative log-likelihood over the dataset: 
\begin{equation}\label{eq:mle}
\mathcal{L}_{\mathrm{MLE}}(\phi;\mathcal{D})
=
-\sum_{i=1}^{m}\sum_{t=1}^{T_i}
\log p_{\phi}\!\left(y_t^{(i)}| x^{(i)}, y_{<t}^{(i)}\right).
\end{equation}
\subsection{Optimization Objective with Role Clarity Regularization}
\label{subsec:rc_obj}
 We now incorporate role clarity as a regularization. 

\textbf{Role‑clarity regularizer.}
Let the $n$ agents' role descriptions as  $\{R_i\}_{i=1}^{n}$.
For each training example, we collect the generated behavior trajectories over $K$ interaction rounds, denoted by $\{B_i^{(k)}\}_{k=1}^{K}$.

Following Section~\ref{subsec:rc_def}, we encode the role descriptions \(\{R_i\}_{i=1}^n\) and the behavior trajectories \(\left\{\{B_i^{(k)}\}_{k=1}^{K}\right\}_{i=1}^n\) into embedding vectors \(\{\mathbf{r}_i(\phi)\}_{i=1}^n\) and \(\{\mathbf{b}_i(\phi)\}_{i=1}^n\), respectively. Based on the embeddings, we obtain the differentiable role clarity matrix \(M(\phi)\) via Equation~(\ref{role}).

Furthermore, we define the role clarity regularizer as the average negative log‑likelihood of the correct role assignments:
\begin{equation}\label{eq:rc_ce}
\mathcal{L}_{\mathrm{RC}}^{\mathrm{CE}}(\phi)
=
-\frac{1}{n}\sum_{i=1}^{n}
\log \text{softmax}_{\tau}(s_{ii}(\phi)).
\end{equation}
Note that Equation~(\ref{eq:rc_ce}) is  derived from the cross-entropy loss $-\sum_{i=1}^ny_i\log(\text{softmax}_{\tau}(s_{ij}(\phi)))$ and $y_i$ is a one‑hot assignment, i.e., $y_i=0$ for $\text{softmax}_{\tau}(s_{ij}(\phi))$ with $i\ne j$.
Minimizing $\mathcal{L}_{\mathrm{RC}}^{\mathrm{CE}}$ encourages role consistency while penalizing role overstepping.

By introducing Equation (\ref{eq:rc_ce}) as a regularizer into Equation (\ref{eq:mle}), we obtain the overall optimization objective
\begin{equation}\label{eq:final_obj}
\mathcal{L}(\phi)
=
\mathcal{L}_{\mathrm{MLE}}(\phi;\mathcal{D})
+\lambda\,\mathcal{L}_{\mathrm{RC}}^{\text{CE}}(\phi).
\end{equation}

\textbf{LoRA fine tuning.}
Optimizing all model parameters using Equation~\eqref{eq:final_obj} in end‑to‑end way is computationally expensive for large language models. To enable parameter‑efficient adaptation while preserving the pretrained backbone, we adopt  LoRA \cite{DBLP:conf/iclr/HuSWALWWC22}, which freezes the original model parameters and learns low‑rank updates to selected weight matrices.

Concretely, for a Transformer weight matrix $\phi_0\in\mathbb{R}^{d_{\text{out}}\times d_{\text{in}}}$, LoRA parameterizes the adapted weight as
\begin{equation}
\phi = \phi_0 + \Delta \phi,\quad \Delta \phi = BA,
\end{equation}
where $A\in\mathbb{R}^{r\times d_{\text{in}}}$ and $B\in\mathbb{R}^{d_{\text{out}}\times r}$ with $r\ll \min(d_{\text{in}},d_{\text{out}})$. During fine tuning, only the low‑rank factors $\{A,B\}$ are trainable, while $\phi_0$ remains frozen, substantially reducing the number of optimized parameters.

This parameterization is fully compatible with the role clarity regularization objective.  Because 
the role embeddings \(\{\mathbf{r}_i(\phi)\}_{i=1}^{n}\) are obtained with the pretrained parameters fixed, whereas the behavior embeddings \(\{\mathbf{b}_i(\phi)\}_{i=1}^{n}\) are obtained with parameters that are updated through training ( see section~\ref{subsec:rc_def}), the resulting role clarity matrix $M(\phi)$ remains differentiable with respect to model activations, then $\mathcal{L}_{\mathrm{RC}}^{\mathrm{CE}}$ is optimizable. Therefore, the loss  Equation~\eqref{eq:final_obj} is minimized jointly over both task supervision and role adherence under the frozen backbone.

In the inference phase, the learned low-rank update can be merged into the base weights (i.e., forming $\phi_0+BA$), avoiding the per‑step adapter overhead in the forward pass.

Algorithm~\ref{alg:lora_role_clarity_single} summarizes the complete training procedure.
\begin{algorithm}[!htb]
  \caption{LoRA Fine Tuning for Agent $A_i$ with Role Clarity Regularization}
  \label{alg:lora_role_clarity_single}
  \begin{algorithmic}[1]
    \STATE {\bfseries Input:} Trajectory dataset $\mathcal{D}=\Big\{\{B_i^{(k)}\}_{k=1}^{K_{m_e}}\Big\}_{m_e=1}^{M_e}$,
with $|\mathcal{D}|=M_e$, collected via rejection sampling;  role descriptions $\{R_j\}_{j=1}^{n}$; pretrained model parameters $\phi_0$; LoRA rank $r$; regularization weight $\lambda$; learning rate $\eta$; training epochs $N$; softmax temperature $\tau$.
    \STATE {\bfseries Initialize:} LoRA parameters $\psi$; Freeze backbone $\phi_0$.
    \COMMENT{Compute hidden states and role embeddings}
    \FOR{$j=1$ {\bfseries to} $n$}
        \STATE $\mathbf{h}_{R_i,1:M}\gets\text{ModelForward}(\phi_0,R_j)$
        \STATE $\mathbf{r}_i(\phi_0) \gets \mathcal{A}(\mathbf{h}_{R_i,1:M})$ \COMMENT{aggregate final‑layer hidden states}
    \ENDFOR
    \FOR{epoch = 1 {\bfseries to} $N$}
     \FOR{sample $m_e=1$ {\bfseries to} $M_e$}
        \COMMENT{Compute hidden states and trajectory embedding}
        \STATE $\mathbf{h}_{B_i,1:T}\gets\text{ModelForward}(\phi_0,\psi,\{B_i^{(k)}\}_{k=1}^{K_{m_e}})$
        \STATE $\mathbf{b}_i^{m_e}(\phi_0,\psi) \gets \mathcal{A}(\mathbf{h}_{\{B_i^{(k)}\}_{k=1}^{K_{m_e}},1:T})$ \COMMENT{aggregate final‑layer hidden states}
        
        
        \COMMENT{Compute similarity scores between behavior and all role embeddings}
        \FOR{$j=1$ {\bfseries to} $n$}
          \STATE $s_{ij}^{m_e} \gets \cos\!\big(\mathbf{b}_i^{m_e}(\phi_0,\psi),\,\mathbf{r}_j(\phi_0)\big)$
        \ENDFOR

        \COMMENT{Softmax normalization with temperature}
        \STATE 
        \[
          \text{softmax}\left(s_{ij}^{m_e}\right) \gets \frac{\exp(s_{ij}^{m_e}/\tau)}{\sum_{j=1}^{n}\exp(s_{ij}^{m_e}/\tau)}
        \]
        
        \COMMENT{Role clarity classification loss (cross‑entropy)}
        \STATE $\mathcal{L}_{\mathrm{RC}}^{m_e} \gets -\log  \text{softmax}\left(s_{ii}^{m_e}\right)$

        \COMMENT{Trajectory generation loss (e.g., next‑token cross‑entropy)}
        \STATE $\mathcal{L}_{\mathrm{traj}}^{m_e} \gets \text{ComputeTrajLoss}(\phi_0,\psi, \{B_i^{(k)}\}_{k=1}^{K_{m_e}})$

        \COMMENT{Total loss with role clarity regularization}
        \STATE $\mathcal{L}^{m_e} \gets \mathcal{L}_{\mathrm{traj}}^{m_e} + \lambda\,\mathcal{L}_{\mathrm{RC}}^{m_e}$

      \COMMENT{Gradient update of LoRA parameters only}
      \STATE $\psi \gets \psi - \eta\,\nabla_{\psi}\mathcal{L}^{m_e}$
      \ENDFOR
    
    \COMMENT{Calculate the role clarity score}
      \STATE $C(M(\phi_0,\psi))=\frac{1}{1+\|M(\phi_0,\psi)\|_F}$

    \ENDFOR

    \STATE {\bfseries Output:} Fine tuned model with merged LoRA parameters applied on backbone $\phi_0$
  \end{algorithmic}
\end{algorithm}
\section{Experiments}
In this section, we conduct experiments in the ChatDev system to evaluate the effectiveness of our proposed role-clarity regularizer in improving (i) role consistency, measured by the role-overstepping rate and role-clarity score, and (ii) end-to-end task performance on SRDD (Software Requirement Description Dataset), evaluated by completeness, executability, consistency, and the composite Quality score.

\subsection{Experimental Setup}
{\bf Benchmarks.}
We use the SWE-Dev benchmark \cite{DBLP:journals/corr/abs-2505-16975}, which simulates realistic software repository environments and development cycles.
Specifically, SWE-Dev contains 14{,}000 training instances and 500 test instances. 
We evaluate downstream end-to-end performance on SRDD, a curated collection of natural-language software requirement descriptions introduced in ChatDev \cite{DBLP:conf/acl/QianLLCDL0CSCXL24}.
SRDD covers software categories from major platforms (e.g., Ubuntu, Google Play, Microsoft Store, and Apple Store) and contains 1{,}200 task prompts across five domains, further divided into 40 subcategories and 
each subcategory contains 30 unique task prompts. \cite{DBLP:journals/corr/abs-2303-17760, DBLP:conf/acl/QianLLCDL0CSCXL24}.

{\bf Backbone Models and Fine Tuning Targets.}
We instantiate ChatDev  system with two  language models: Qwen2.5-7B-Instruct \cite{DBLP:journals/corr/abs-2412-15115} and Llama-3.1-8B-Instruct \cite{DBLP:journals/corr/abs-2302-13971}. 
Role overstepping in ChatDev primarily occurs during the requirements analysis stage between CEO and CPO agents. Therefore, we only fine tune  the CEO and CPO agents.

{\bf Parameter-efficient Fine Tuning.}
For parameter-efficient adaptation of the open-weight model (Qwen2.5-7B-Instruct and Llama-3.1-8B-Instruct), we apply LoRA adapters to the query, key, value, and output projection matrices of each self-attention sublayer. The LoRA configuration uses rank $r=16$, scaling factor $\alpha=16$, and a dropout rate of 0.05.

{\bf Optimization and Model Selection.}
We fine tune with a learning rate of $5\times10^{-5}$ and a batch of 32 sequences per gradient update, for up to 10 epochs. 
Checkpoints are saved every 50 steps, and the best checkpoint is selected based on validation loss on a held-out set of 100 examples. 
Fine-tuning is initialized from the respective pretrained weights of Qwen2.5-7B-Instruct and Llama-3.1-8B-Instruct.
\subsection{Trajectory Collection}
We use Qwen2.5-7B-Instruct to collect training data for the ChatDev as the following:

(1) \textbf{Multi‑turn ChatDev Inference.} For each training instance, the ChatDev system executes multi‑turn reasoning to generate a candidate interaction trajectory comprising messages exchanged among agents.\\
(2) \textbf{Rejection Sampling.} During each iteration, we apply rejection sampling based on  the termination token \texttt{<INFO>}. We only retain those trajectories for which both CPO and CEO agents adhere to their role specifications and formatting requirements.\\
(3) \textbf{Role Consistency Fine tuning.} The retained trajectories are used to construct fine tuning datasets for the CPO and CEO agents. 
\subsection{Evaluation Metrics}
We evaluate our method from two perspectives: role consistency and end-to-end task performance.

\textbf{Role consistency.}
We use the role clarity score in Equation~(\ref{2.1}) and the role overstepping rate
\[
\mathrm{RoleOverstepRate}=\frac{\#\text{ overstepping test cases}}{\#\text{ total test cases}}
\]
to evaluate role consistency, where an \emph{overstepping test case} is that the agent violates its assigned role boundary at least once during the $10$-turn interactions.

\textbf{End-to-end task performance (SRDD).}
We use the following  four fundamental dimensions to evaluate  SRDD\cite{DBLP:journals/corr/abs-2406-08979}.\\
$\bullet$ \textbf{Completeness} ($\alpha \in [0,1]$): the proportion of generated code that is free from ``TODO''-like placeholders, reflecting the degree of code completion.\\
$\bullet$ \textbf{Executability} ($\beta \in [0,1]$): the proportion of generated software projects that compile without errors (and are ready to run) in the target environment.\\
$\bullet$ \textbf{Consistency} ($\gamma \in [0,1]$): the semantic alignment between the generated software and the natural-language requirements, computed via embedding similarity.\\
$\bullet$ \textbf{Quality} ($q \in [0,1]$): an overall indicator aggregating the above dimensions, i.e., $ q=\frac{\alpha+\beta+\gamma}{3}.$
\subsection{Experimental Results}
\subsubsection{Role Consistency on SWE-Dev}
In this part, we report the role overstepping rates and role clarity scores of Qwen2.5-7B-Instruct and Llama-3.1-8B-Instruct on the SWE-Dev evaluation set under four configurations: (i) Base–Base (no fine-tuning for either agent), (ii) fine-tuning only the CPO agent, (iii) fine-tuning only the CEO agent, and (iv) jointly fine-tuning both agents (FT) with role-clarity regularization. We further include an ablation setting that removes the role-clarity regularization term while keeping all other training conditions identical. Note that \texttt{<INFO>} denotes a correctly formatted decision output, whereas \texttt{INFO} denotes an incorrectly formatted output that still reflects the intended decision, $\Delta$ indicates the change relative to the Base–Base baseline.
\begin{table*}[h]
  \centering
 \caption{
Role  overstepping rate (\%)  on the SWE-Dev evaluation set for the ChatDev system with Qwen as the backbone (with role clarity regularization). Results are reported for the easy and hard subsets (250 cases each) and for the full set (500 cases; total).
}
  \label{Table 1}
  \begin{tabular}{c c c c c c c c}
    \toprule
    \multirow{2}{*} {CEO} &  \multirow{2}{*} {CPO}  &  \multicolumn{4}{c}{Overstepping Rate (\texttt{<INFO>/\texttt{INFO)}}} & \multicolumn{2}{c}{Total Overstepping Rate(\texttt{<INFO>/INFO})} \\
    \cmidrule(lr){3-6}  \cmidrule(lr){7-8}
     & & easy &$\Delta(\%)$& hard &$\Delta(\%)$ &total & $\Delta(\%)$\\
     \midrule
     Base & Base & 43.0/65.8&--&  49.7/68.2& --& 46.4/67.0& -- \\
      \midrule
     FT & Base & 37.2/37.6&-5.8/-28.2 &44.0/44.0 &-5.7/-24.2 & 40.6/40.8&-5.8/-26.2  \\
      \midrule
     Base & FT &19.2/34.4 &-23.8/-31.4 &16.8/30.4 &-32.9/-37.8 &18.0/32.4 &-28.4/-34.6 \\
      \midrule
     FT& FT & \textbf{10.0/10.4}& \textbf{-33.0/-55.4}& \textbf{6.8/7.6}& \textbf{-42.9/-60.6}& \textbf{8.4/9.0}&\textbf{-38.0/-58.0}\\
    \bottomrule
  \end{tabular}
\end{table*}

\begin{table*}[h]
  \centering
 \caption{
Role  overstepping rate (\%)  on the SWE-Dev evaluation set for the ChatDev system with Llama as the backbone (with role clarity regularization). Results are reported for the easy and hard subsets (250 cases each) and for the full set (500 cases; total).
}
  \label{Table 3}
  \begin{tabular}{c c c c c c c c}
    \toprule
    \multirow{2}{*} {CEO} &  \multirow{2}{*} {CPO}  &  \multicolumn{4}{c}{Overstepping Rate (\texttt{<INFO>/\texttt{INFO)}}} & \multicolumn{2}{c}{Total Overstepping Rate(\texttt{<INFO>/INFO})} \\
    \cmidrule(lr){3-6}  \cmidrule(lr){7-8}
     & & easy &$\Delta(\%)$& hard &$\Delta(\%)$ &total & $\Delta(\%)$\\
     \midrule
     Base & Base & 45.2/46.4&--&  41.6/42.8& --& 43.4/44.6& -- \\
      \midrule
     FT & Base & 0.8/0.8& -44.4/-45.6&0.5/0.5 &-41.1/-42.3 & 0.65/0.65& -42.75/-43.95\\
      \midrule
     Base & FT &16.8/16.8 &-28.4/-29.6&14.4/15.6 & -27.2/-27.2&15.6/16.2&-27.8/-28.4 \\
      \midrule
     FT& FT & \textbf{0.4/0.4}& \textbf{-44.8/-46.0}& \textbf{0.0/0.0}&\textbf{-41.6/-42.8} & \textbf{0.2/0.2}&\textbf{-43.2/-44.4}\\
    \bottomrule
  \end{tabular}
\end{table*}
\begin{table*}[!ht]
  \centering
 \caption{Role clarity score on the SWE-Dev evaluation set for the ChatDev system with Qwen as the backbone (with role clarity regularization). Results are reported for the easy and hard subsets (250 cases each) and for the full set (500 cases; total).}
  \label{Table 2}
  \begin{tabular}{c c c c c c c c}
    \toprule
    \multirow{2}{*} {CEO} &  \multirow{2}{*} {CPO}  &  \multicolumn{4}{c}{Role clarity score} & \multicolumn{2}{c}{Total role clarity score} \\
    \cmidrule(lr){3-6}  \cmidrule(lr){7-8}
     & & easy &$\Delta$& hard &$\Delta$ &total & $\Delta$\\
     \midrule
     Base & Base & 0.5360&--& 0.5295& --& 0.5328& -- \\
      \midrule
     FT & Base &0.6583&+0.1223 &0.6440 &+0.1145 & 0.6511& 0.1183\\
      \midrule
     Base & FT &0.6014 &+0.0654 &0.6044 &+0.0749&0.6029 &+0.0701 \\
      \midrule
     FT& FT &\textbf{0.9081} & \textbf{+0.3721}&\textbf{0.9076} & \textbf{+0.3781}&\textbf{0.9079} &\textbf{+0.3751}\\
    \bottomrule
  \end{tabular}
\end{table*}
\begin{table*}[!ht]
  \centering
 \caption{Role clarity score on the SWE-Dev evaluation set for the ChatDev system with Llama as the backbone (with role clarity regularization). Results are reported for the easy and hard subsets (250 cases each) and for the full set (500 cases; total).}
  \label{Table 4}
  \begin{tabular}{c c c c c c c c}
    \toprule
    \multirow{2}{*} {CEO} &  \multirow{2}{*} {CPO}  &  \multicolumn{4}{c}{Role clarity score} & \multicolumn{2}{c}{Total role clarity score} \\
    \cmidrule(lr){3-6}  \cmidrule(lr){7-8}
     & & easy &$\Delta$& hard &$\Delta$ &total & $\Delta$\\
     \midrule
     Base & Base & 0.5000&--& 0.5012& --& 0.5007& -- \\
      \midrule
     FT & Base &0.5975&+0.0975 &0.5496 & +0.0484& 0.5735& +0.0728\\
      \midrule
     Base & FT &0.5635 & +0.0635&0.5655 &+0.0643&0.5645 &+0.0638 \\
      \midrule
     FT& FT &\textbf{0.8525} & \textbf{+0.3525}&\textbf{0.8535} & \textbf{+0.3523}&\textbf{0.8530} &\textbf{+0.3523}\\
    \bottomrule
  \end{tabular}
\end{table*}
\begin{table*}[!ht]
  \centering
 \caption{
Role overreach rate (\%) on the SWE-Dev evaluation set for the ChatDev system with Qwen as the backbone, with role clarity regularization disabled. Results are reported for the easy and hard subsets (250 cases each) and for the full set (500 cases; total).
}
  \label{Table 5}
  \begin{tabular}{c c c c c c c c}
    \toprule
    \multirow{2}{*} {CEO} &  \multirow{2}{*} {CPO}  &  \multicolumn{4}{c}{Overstepping Rate (\texttt{<INFO>/\texttt{INFO)}}} & \multicolumn{2}{c}{Total Overstepping Rate(\texttt{<INFO>/INFO})} \\
    \cmidrule(lr){3-6}  \cmidrule(lr){7-8}
     & & easy &$\Delta(\%)$& hard &$\Delta(\%)$ &total & $\Delta(\%)$\\
     \midrule
     Base & Base & 43.0/65.8&--&  49.7/68.2& --& 46.4/67.0& -- \\
      \midrule
     FT & Base & 36.4/36.4&-6.6/-29.4 &45.6/46.0 &-4.1/-22.2&41.0/41.2&-5.4/-25.8\\
      \midrule
     Base & FT &24.4/37.2 &-18.8/-28.8 &21.6/34.4 &-28.1/-33.8&23.0/35.8 &-23.4/-31.2 \\
      \midrule
     FT& FT & \textbf{12.0/12.0}& \textbf{-31.0/-53.8}&\textbf{9.2/9.2}&\textbf{-40.5/-59.0}& \textbf{10.6/10.6}&\textbf{-35.8/-56.4}\\
    \bottomrule
  \end{tabular}
\end{table*}
\begin{table*}[!ht]
  \centering
 \caption{
Role overreach rate (\%) on the SWE-Dev evaluation set for the ChatDev system with Llama as the backbone, with role clarity regularization disabled. Results are reported for the easy and hard subsets (250 cases each) and for the full set (500 cases; total).
}
  \label{Table 7} 
  \begin{tabular}{c c c c c c c c}
    \toprule
    \multirow{2}{*} {CEO} &  \multirow{2}{*} {CPO}  &  \multicolumn{4}{c}{Overstepping Rate (\texttt{<INFO>/\texttt{INFO)}}} & \multicolumn{2}{c}{Total Overstepping Rate(\texttt{<INFO>/INFO})} \\
    \cmidrule(lr){3-6}  \cmidrule(lr){7-8}
     & & easy &$\Delta(\%)$& hard &$\Delta(\%)$ &total & $\Delta(\%)$\\
     \midrule
     Base & Base & 45.2/46.4&--& 41.6/42.8& --& 43.4/44.6& -- \\
      \midrule
     FT & Base &1.0/1.0 & -44.2/-45.4&0.8/0.8 &-40.8/-42.0&0.9/0.9&-42.5/-43.7\\
      \midrule
     Base & FT &16.4/16.8 & -28.8/-29.6&16.0/16.0 &-25.6/-26.8&16.2/16.4 & -27.2/-28.2\\
      \midrule
     FT& FT & \textbf{0.0/0.0}& \textbf{-45.2/-46.4}&\textbf{0.0/0.0}&\textbf{-41.6/-42.8}& \textbf{0.0/0.0}&\textbf{-43.4/-44.6}\\
    \bottomrule
  \end{tabular}
\end{table*}
\begin{table*}[!ht] 
  \centering
 \caption{Role clarity score (\%) on the SWE-Dev evaluation set for the ChatDev system with Qwen as the backbone, with role clarity regularization disabled. Results are reported for the easy and hard subsets (250 cases each) and for the full set (500 cases; total).}
  \label{Table 6}
  \begin{tabular}{c c c c c c c c}
    \toprule
    \multirow{2}{*} {CEO} &  \multirow{2}{*} {CPO}  &  \multicolumn{4}{c}{Role clarity score} & \multicolumn{2}{c}{Total role clarity score} \\
    \cmidrule(lr){3-6}  \cmidrule(lr){7-8}
     & & easy &$\Delta$& hard &$\Delta$ &total & $\Delta$\\
     \midrule
     Base & Base & 0.5360&--& 0.5295& --& 0.5328& -- \\
      \midrule
     FT & Base &\textbf{0.5694}&\textbf{+0.0334}&\textbf{0.5622}&\textbf{+0.0327}& \textbf{0.5658}&\textbf{ 0.0330} \\
      \midrule
     Base & FT &0.5086 &-0.0274 &0.5067 &-0.0219 &0.5077 &-0.0251 \\
      \midrule
     FT& FT &0.5208 & -0.0152&0.5208& -0.0087&0.5208&-0.012\\
    \bottomrule
  \end{tabular}
\end{table*}
\begin{table*}[!ht] 
  \centering
 \caption{Role clarity score (\%) on the SWE-Dev evaluation set for the ChatDev system with Llama as the backbone, with role clarity regularization disabled. Results are reported for the easy and hard subsets (250 cases each) and for the full set (500 cases; total).}
  \label{Table 8}
  \begin{tabular}{c c c c c c c c}
    \toprule
    \multirow{2}{*} {CEO} &  \multirow{2}{*} {CPO}  &  \multicolumn{4}{c}{Role clarity score} & \multicolumn{2}{c}{Total role clarity score} \\
    \cmidrule(lr){3-6}  \cmidrule(lr){7-8}
     & & easy &$\Delta$& hard &$\Delta$ &total & $\Delta$\\
     \midrule
     Base & Base & 0.5000&--& 0.5012& --& 0.5007& -- \\
      \midrule
     FT & Base &\textbf{0.5579}&\textbf{+0.0579}&\textbf{0.5549}&\textbf{+0.0537}& \textbf{0.5564}&\textbf{+0.0557} \\
      \midrule
     Base & FT &0.5008 &0.0008&0.4975 & -0.0037&0.4992 &-0.0015\\
      \midrule
     FT& FT &0.5305 &+0.0305 &0.5287& 
     +0.0275&0.5296&+0.0289\\
    \bottomrule
  \end{tabular}
\end{table*}
\begin{table*}[!ht]
  \centering
 \caption{SRDD end-to-end performance and role-consistency for Qwen and Llama. ``FT'' denotes fine-tuning both the CPO and CEO agents on SWE-Dev; non-FT uses the original base model. Lower is better for overstepping rate, and higher is better for the other metrics.}
  \label{Table end}
  \begin{tabular}{c c c c c c c c}
    \toprule
     & Completeness & Executability &Consistency& Quality& Overstepping& clarity score\\
      \midrule
     Qwen  &0.7002  &\textbf{0.9713}&0.3593 &0.6769 &0.8133&0.5278\\
       Qwen (FT) & \textbf{0.7272}$^{\uparrow 2.70\%}$ &0.9561$^{\downarrow 1.52\%}$&\textbf{0.3894}$^{\uparrow 3.01\%}$ &\textbf{0.6909}$^{\uparrow 1.40\%}$ &\textbf{0.1554}$^{\downarrow 65.79\%}$&\textbf{0.8817}$^{\uparrow 35.39\%}$\\
      \midrule
     Llama& 0.7106&0.8243 &0.3218&0.6174&0.6225& 0.5003 \\
        Llama (FT) &\textbf{0.7635}$^{\uparrow 5.29\%}$ &\textbf{0.8716}$^{\uparrow 4.73\%}$ &\textbf{0.3937}$^{\uparrow 7.19\%}$&\textbf{0.6763}$^{\uparrow 5.89\%}$&\textbf{0}$^{\downarrow 62.25\%}$&\textbf{0.8431}$^{\uparrow 34.28\%}$ \\
    \bottomrule
  \end{tabular}
\end{table*}

\textbf{Fine-tuning with Role Clarity Regularization.}
Tables ~\ref{Table 1}-~\ref{Table 4} show that fine-tuning with role-clarity regularization consistently reduces the role overstepping rate and improves the role clarity score on both the easy and hard subsets. 
Moreover, this approach strengthens format compliance: the overstepping rates computed under the strict formatting criterion (\texttt{<INFO>}) and the relaxed criterion (\texttt{INFO}) differ by at most  0.6\%, indicating that the model more reliably produces standardized outputs while preserving consistent decision intent.\\
\textbf{Fine-tuning without Role Clarity Regularization.}
Tables~\ref{Table 5}-~\ref{Table 8} show that, when fine tuning is performed without the role-clarity regularization term, the role overstepping rate decreases, but the role clarity score is almost the same as the Base-Base baseline, even shows a slight downward trend.

\subsubsection{End-to-End Task Performance on SRDD}
Table~\ref{Table end} shows that after fine-tuning both the CPO and CEO agents on SWE-Dev, effective improvements are observed in both end-to-end task performance and role consistency on SRDD. Specifically, across the end-to-end evaluation dimensions, the fine-tuned Qwen and Llama consistently improve task performance, indicating better alignment with requirements and more complete software generation.
Meanwhile, role-consistency metrics also improve markedly during SRDD inference: the role-overstepping rate is substantially reduced and the role-clarity score increases, suggesting tighter adherence to role boundaries under deployment.
Overall, these results support the claim that our role clarity regularizer provides a structural mechanism to enhance the reliability and  stability of multi-agent systems.
\section{Conclusion}
In this paper, we  propose a quantifiable role clarity to improve the role consistency in multi-agent systems. To our best knowledge, this is the first proposed. Experiments conducted within the ChatDev framework validate the effectiveness and generalizability of our approach. 
Importantly, transferring the fine-tuned models to the SRDD task yields consistent improvements in end-to-end task performance, demonstrating that our role consistency objective provides measurable downstream benefits.
These findings support the view that structurally grounded role-clarity supervision can improve the robustness of LLM-based multi-agent systems and facilitate deployment in task-oriented applications. 

We anticipate that our work inspire a paradigm shift in multi‑agent role management, from qualitative prompting to quantitative optimization. 
In future work, we will (1) extend this approach to broader multi-agent settings (e.g., tool use and long-horizon planning); and (2) develop standardized benchmarks with automated evaluation for role consistency.

\bibliographystyle{unsrtnat}
\bibliography{references}

\end{document}